\documentclass[journal]{IEEEtran}
\ifCLASSINFOpdf
\else
\fi
\usepackage{graphicx}
\usepackage{amsmath}  
\usepackage{mathtools}
\usepackage{algorithm}
\usepackage[noend]{algpseudocode}
\usepackage{booktabs}
\usepackage{subcaption}
\usepackage{multicol}
\usepackage{framed} 
\usepackage[framed]{ntheorem}
\newframedtheorem{frm-thm}{Premise}
\newframedtheorem{frm-lmm}{Lemma}
\newframedtheorem{frm-rule}{``Condensed'' Rule}
\usepackage{amssymb}
\usepackage{array,multirow}

\usepackage[table]{xcolor}
\usepackage{balance}

\begin{document}
%
\title{Review, Analysis and Design of a Comprehensive Deep Reinforcement Learning Framework}
%
%
%

\author{Ngoc~Duy~Nguyen,
        Thanh~Thi~Nguyen,
        Hai~Nguyen,
				Doug~Creighton,
        Saeid~Nahavandi
\thanks{
Ngoc Duy Nguyen, Doug Creighton and Saeid Nahavandi are with the Institute for Intelligent Systems Research and Innovation, Deakin University, Waurn Ponds Campus, Geelong, Victoria, Australia (e-mails: duy.nguyen@deakin.edu.au, douglas.creighton@deakin.edu.au and saeid.nahavandi@deakin.edu.au).}
\thanks{
Thanh Thi Nguyen is with the School of Information Technology, Deakin University, Burwood Campus, Melbourne, Victoria, Australia (e-mail: thanh.nguyen@deakin.edu.au).}
\thanks{
Hai Nguyen is with Khoury College of Computer Science, Notheastern University, Boston, USA (e-mail: hainguyen@ccs.neu.edu)
}
}
%
%

\markboth{}%
{Shell \MakeLowercase{\textit{et al.}}: Review, Analysis and Design of a Comprehensive Deep Reinforcement Learning Framework}
%



\maketitle

\begin{abstract}
The integration of deep learning to reinforcement learning (RL) has enabled RL to perform efficiently in high-dimensional environments. Deep RL methods have been applied to solve many complex real-world problems in recent years. However, development of a deep RL-based system is challenging because of various issues such as the selection of a suitable deep RL algorithm, its network configuration, training time, training methods, and so on. This paper proposes a comprehensive software framework that not only plays a vital role in designing a connect-the-dots deep RL architecture but also provides a guideline to develop a realistic RL application in a short time span. We have designed and developed a deep RL-based software framework that strictly ensures flexibility, robustness, and scalability. By inheriting the proposed architecture, software managers can foresee any challenges when designing a deep RL-based system. As a result, they can expedite the design process and actively control every stage of software development, which is especially critical in agile development environments. To enforce generalization, the proposed architecture does not depend on a specific RL algorithm, a network configuration, the number of agents, or the type of agents. Using our framework, software developers can develop and integrate new RL algorithms or new types of agents, and can flexibly change network configuration or the number of agents.
\end{abstract}

\begin{IEEEkeywords}
reinforcement learning, deep learning, software architecture, learning systems, multi-agent systems, human-machine interactions, framework.
\end{IEEEkeywords}

%
\IEEEpeerreviewmaketitle

\section{Introduction}
\label{sec:1}
%
%
%
%
\IEEEPARstart{R}{ecent} development of deep learning has been applied to solve various complex real-world problems. Notably, its integration into reinforcement learning (RL) has attracted a great deal of research attention. RL conducts a learning procedure by allowing agents to directly interact with the environment. An RL agent can imitate human learning process to achieve a designated goal, \emph{i.e.}, the agent conducts trial-and-error learning (exploration) and draws on ``experience" (exploitation) to improve its behaviors \cite{1,2}. Therefore, RL is used in countless domains, such as IT resources management \cite{3}, cyber-security \cite{4}, robotics \cite{5, 6, 6b, 7}, surgical robotics \cite{7b, 7c}, control systems \cite{8, 9}, recommendation systems \cite{10}, bidding and advertising campaigns \cite{11}, and video games \cite{12, 13, 14, 14b, 14c}. However, traditional RL methods and dynamic programming \cite{15}, which use a \emph{bootstrapping} mechanism to approximate the objective function, cease to work in high-dimensional environments due to memory and processing power limitations. This so-called issue, \emph{the curse of dimensionality}, creates a major challenge in RL literature. 

\begin{figure}[!b]
\centering
\includegraphics[width=0.95\linewidth]{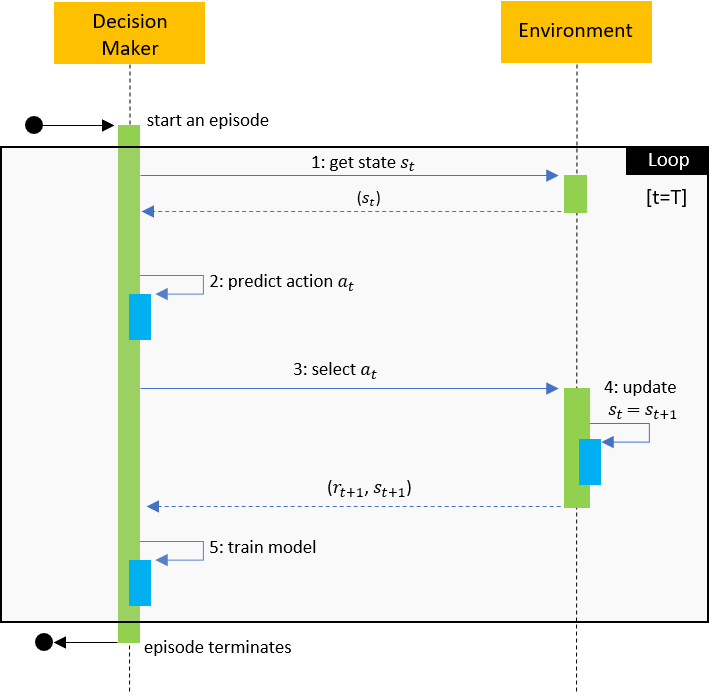}
\caption{Using a UML sequential diagram to describe an RL problem.}
\label{fig:1} 
\end{figure}

Fig.~\ref{fig:1} describes an RL problem by using a \emph{Unified Modeling Language} (UML) \cite{16} sequential diagram. Specifically, the problem includes two entities: a decision maker and the environment. The environment can be an artificial simulator or a wrapper of the real-world environment. While the environment is a \emph{passive} entity, the decision maker is an \emph{active} entity that periodically interacts with the environment. In the RL context, a decision maker and an agent can be interchangeable, though they can be two identified objects from a software design perspective. 

At first, the decision maker perceives a state $s_t$ from the environment. Then it uses its internal model to select the corresponding action $a_t$. The environment interacts with the chosen action $a_t$ by sending a numerical reward $r_{t+1}$ to the decision maker. The environment also brings the decision maker to a new state $s_{t+1}$. Finally, the decision maker uses the current transition $\vartheta=\{a_t, s_t, r_{t+1}, s_{t+1}\}$ to update its decision model. This process is iterated until $t$ equals $T$, where $s_T$ denotes the terminal state of an episode. There are different methods to develop a decision model, such as fuzzy logic \cite{17}, genetic algorithms \cite{18, 18b}, or dynamic programming \cite{19}. In this paper, however, we consider a deep neural network as the decision model.

The previous diagram infers that RL is \emph{online} learning because the model is updated with incoming data. However, RL can be performed offline via a \emph{batch learning} \cite{20} technique. In particular, the current transition $\vartheta$ can be stored in an \emph{experience replay} \cite{21} and retrieved later to train the decision model. Finally, the goal of an RL problem is to maximize the expected sum of discounted reward $R_t$, \emph{i.e.},

\begin{equation*}
R_t = r_{t+1} + \gamma r_{t+2} + \gamma^2 r_{t+3} + ... + \gamma^{T-t-1} r_T, 
\end{equation*} 

\noindent
where $\gamma$ denotes the discounted factor and $0 < \gamma \leq 1$. 

In 2015, Google DeepMind \cite{22} announced a breakthrough in RL by combining it with deep learning to create an intelligent agent that can beat a professional human player in a series of 49 Atari games. The idea was to use a deep neural network with convolutional layers \cite{23} to directly process raw images (states) of the game screen to estimate subsequent actions. The study is highly valued because it opened a new era of RL with deep learning, which partially solves the curse of dimensionality. In other words, deep learning is a great complement for RL in a wide range of complicated applications. For instance, Google DeepMind created a program, AlphaGo, which beat the Go grandmaster, Lee Sedol, in the best-of-five tournament in 2016 \cite{24}. AlphaGo is a full-of-tech AI that is based on Monte Carlo Tree Search \cite{25}, a hybrid network (policy network and value network), and a self-taught training strategy \cite{26}. Other applications of deep RL can be found in self-driving cars \cite{27, 27b}, helicopters \cite{28}, or even NP-hard problems such as \emph{Vehicle Routing Problem} \cite{29} and combinatorial graph optimization \cite{30}.

As stated above, deep RL is crucial owing to its appealing learning mechanism and widespread applications in the real world. In this study, we further delve into practical aspects of deep RL by analyzing challenges and solutions while designing a deep RL-based system. Furthermore, we consider a real-world scenario where multiple agents, multiple objectives, and human-machine interactions are involved. Firstly, if we can take advantage of using multiple agents to accomplish a designated task, we can shorten the \emph{wall time}, \emph{i.e.}, the computational time to execute the assigned task. Depending on the task, the agents can be \emph{cooperative} or \emph{competitive}. In the cooperative mode, agents work in \emph{parallel} or in \emph{pipeline} to achieve the task \cite{31}.  In the case of competition, agents are scrambled, which basically raises the \emph{resource hunting} problem \cite{32}. However, in contrast to our imagination, competitive learning can be fruitful. Specifically, the agent is trained continually to place the opponent into a disadvantaged position and the agent is improved over time. Because the opponent is also improved over time, this phenomenon eventually results in the \emph{Nash equilibrium} \cite{33}. Moreover, competitive learning originates the self-taught strategy (\emph{e.g.} AlphaGo) and a series of techniques such as the \emph{Actor-Critic architecture} \cite{34, 35}, opponent modeling \cite{36, 37}, and \emph{Generative Adversarial Networks} \cite{38}. Finally, we notice the problem of \emph{moving target} \cite{38b, 38c} in multi-agent systems, which describes a scenario when the decision of an agent depends on other agents, thus the optimal policy becomes \emph{non-stationary} \cite{38c, 38d}.

Secondly, a real-world objective is often complicated as it normally comprises of multiple sub-goals. It is straightforward if sub-goals are \emph{non-conflicting} because they can be seen as a single composite objective. The more difficult case is when there are \emph{conflicting objectives}. One solution is to convert a multi-objective problem into a single objective counterpart by applying \emph{scalarization} through an application of a linear weighted sum for individual objective \cite{39} or non-linear methods \cite{40}. These approaches are categorized as \emph{single-policy methods}. In contrast, the \emph{multi-policy methods} \cite{41} seek multiple optimal policies at the same time. Although the number of multi-policy methods is restricted, it can be powerful. For instance, the \emph{Convex Hull Value Iteration} algorithm \cite{41b} computes a set of objective combinations to retrieve all deterministic optimal policies. Finally, to benchmark a multi-objective method, we find or approximate a boundary surface, namely \emph{Pareto dominance}, which presents the maximum performance of different weights (if scalarization is used) \cite{42}. Recent studies have tried to integrate multi-objective mechanisms into deep RL \cite{43, 44, 45}.

\begin{table*}[!t]
\renewcommand{\arraystretch}{1.3}
\caption{Key RL terminologies}
\label{table:1}
\centering
\begin{tabular}{|c|c|c|c|}
\hline
\hline
\textbf{Term} & \textbf{Description} & \textbf{Pros} & \textbf{Cons} \\
\hline
\multirow{4}{*}{Model-free RL} &The environment is a black box. Agents mostly&The algorithm does not &Requires a large amount of \\
                         & conduct a trial-and-error procedure to learn on its own.&need a model of the&samples\\
                         & They use rewards to update their decision models&environment&\\
                         
\hline
\multirow{5}{*}{Model-based RL} &Agents construct a model that&Speed up learning&Having an accurate\\
 					      & simulates the environment and use it to generate&and improve sample& and useful model is often \\
 					      & future episodes. By using the model, agents can&efficiency& challenging\\
 					      & estimate not only actions but also future states&&\\
 					      
\hline
\hline
\multirow{3}{*}{Temporal difference learning}&Use TD error to estimate the value function.&Fast convergence as it&Estimates can be\\
  & For example, in Q-learning, &does not need to wait until &biased\\
                             
                                      &$Q(s_t,a_t) = Q(s_t,a_t) + \beta(r_t + \gamma \max_a Q(s_{t+1}, a))$ &the episode ends&\\
\hline
\multirow{3}{*}{Monte-Carlo method}&Estimate the value function by obtaining the average &The values are non-biased&Slow convergence and the\\
                                   & of the same values in different episodes &estimates&estimates have high\\
                                   &$Q(s_t,a_t) = \lim_{N->\infty}\sum_{i=1}^N Q(s_t^i,a_t^i) $&&variances. Has to wait until\\
                                   &&&episode ends to do updates\\
\hline
\hline
\multirow{2}{*}{Continuous action space}&The number of control actions is continuous&A policy-based method&Cannot use a\\
					   &&can be used&value-based method\\
\hline			 
\multirow{3}{*}{Discrete action space}&The number of control actions is discrete&Both policy-based method&Intractable if the number\\
					 &and finite& and value-based method&of actions is large\\
					 &&can be used&\\
\hline
\hline
\multirow{2}{*}{Deterministic policy} &The policy maps each state to a specific action&Reduce data sampling&Vulnerable to noise and\\
&&&stochastic environments\\
\hline
\multirow{2}{*}{Stochastic policy}&The policy maps each state to a probability distribution&Better exploration&Requires a large amount of\\
				  &over actions&&samples\\
\hline
\hline
\multirow{2}{*}{On-policy method}&Improve the current policy that the agent&Safer to explore&May be stuck in local \\
&is using to make decisions&&minimum solutions\\
\hline
\multirow{2}{*}{Off-policy method}&Learn the optimal policy (while samples are generated&Instability. Often used with&Might be unsafe because\\
						    &by the behavior policy)&an experience replay&the agent is free to explore\\
\hline
\hline
Fully observable &All agents can observe the complete states of the&Easier to solve than&The number of state\\
environment       &environment&partially observable&can be large\\
&&environments&\\
\hline
Partially observable&Each agent only observes a limited observation&More practical in &More difficult to solve\\
environment          &of the environment&real-world applications&as the agents require\\
&&&to remember past states\\
\hline
\end{tabular}
\end{table*}

Last but not least, human-machine interaction is another key factor in a deep RL-based system. A self-driving car, for example, should accept human intervention in emergency cases \cite{45b, 45c}. Therefore, it is critical to ensure a certain level of safety while designing a hybrid system in which humans and machines can work together. Due to its importance, Google DeepMind and OpenAI have presented novel ways that have encouraged a number of inventions in recent years \cite{46}. For instance, Christiano \emph{et al.} \cite{47} propose a novel scheme that accepts human feedback during the training process. However, the method requires an operator to constantly observe the agent's behavior, which is an onerous and error-prone task. Recent work \cite{48} provides a more practical approach by introducing a behavioral control system. The system is used to control multiple agents in real time via human dictation. Table \ref{table:1} summarizes key terminologies that are widely used in RL contexts.

In summary, the study contributes the following key factors:

\begin{itemize}

\item The paper presents an overall picture of contemporary deep RL. We briefly overview state-of-the-art deep RL methods considering three key factors of a real-world application such as multi-agent learning, multi-objective problems, and human-machine interactions. Thereafter, the paper offers a checklist for software managers, a guideline for software designers, and a technical document for software programmers. 

\item We analyze challenges and difficulties while designing a deep RL-based system and hence mitigate possible mistakes during the development process. In other words, software designers can inherit the proposed design, foresee difficulties, and eventually expedite the entire development procedure, especially in agile software development.

\item Finally, the source code of the proposed framework can be found in \cite{109}. Based on this template, RL beginners can prototype an RL method and develop an RL-based application in a short time span. As a result, the paper is a key factor to share deep RL to a wider community.

\end{itemize}

The paper has the following sections. Section \ref{sec:2} conducts a brief survey of state-of-the-art deep RL methods in different research directions. Section \ref{sec:3} presents our proposed system architecture, which supports multiple agents, multiple objectives, and human-machine interactions. Finally, we conclude the paper in Section \ref{sec:4}.

\section{Literature Review}
\label{sec:2}

\begin{table*}[!t]
\renewcommand{\arraystretch}{1.3}
\caption{Key deep RL methods in literature}
\label{table:2}
\centering
\begin{tabular}{ccccc}
\hline
\hline
\textbf{Method} & \textbf{Description and Advantage} & \textbf{Technical Requirements} & \textbf{Drawbacks} & \textbf{Implementation} \\
\hline
\multicolumn{5}{c}{\cellcolor{black}{\textcolor{white}{Value-based method}}} \\
\hline
DQN & Use a deep convolutional network to & $\bullet$ Experience replay& $\circ$ Excessive memory usage & \cite{65g}\\
         & directly process raw graphical data and & $\bullet$ Target network     & $\circ$ Learning instability& \cite{65h}\\
         &  approximate the action-value function                                 &                               $\bullet$ Q-learning&$\circ$ Only for discrete action&\cite{65j}\\
         &&&space&\cite{65k}\\
\hline
Double DQN& Mitigate the DQN's maximization bias &$\bullet$ Double Q-learning&$\circ$ Inherit DQN's drawbacks&\cite{65k}\\
			       & problem by using two separate networks:       &&&\\
				   & one for estimating the value, one for &&&\\
				   & selecting action.&&&\\
\hline
Prioritized 	          & Prioritize important transitions so          & $\bullet$Importance sampling&$\circ$ Inherit DQN's drawbacks&\cite{65g}\\
Experience Replay & that they are sampled more frequently. &  &$\circ$ Slower than non-prioritized&\cite{65k}\\
                                & Improve sample efficiency & &experience replay (speed)&\\
\hline
Dueling Network& Separate the DQN architecture into&$\bullet$ Dueling network&$\circ$ Inherit DQN's drawbacks&\cite{65k}\\
					     & two streams: one estimates state-value &architecture&&\\
					     & function and one estimates the advantage &$\bullet$ Prioritized replay&&\\
					     & of each action&&&\\
\hline
Recurrent DQN & Integrate recurrency into DQN      &$\bullet$ Long Short Term&$\circ$ Inherit DQN's drawbacks&\cite{65k}\\
				        & Extend the use of DQN in          	  &Memory&&\\
				        & partially observable environments&&&\\
\hline
Attention Recurrent&Highlight important regions of the&$\bullet$ Attention mechanism&$\circ$ Inherit DQN's drawbacks&\cite{54}\\
DQN                         &environment during the training process&$\bullet$ Soft attention&&\\
							  &         &$\bullet$ Hard attention&&\\
\hline
Rainbow &Combine different techniques in DQN&$\bullet$ Double Q-learning&$\circ$ Inherit DQN's drawbacks&\cite{65g}\\
               &variants to provide the state-of-the-art&$\bullet$ Prioritized replay&&\\
               &performance on Atari domain&$\bullet$Dueling network&&\\
               &&$\bullet$Multi-step learning&&\\
               &&$\bullet$Distributional RL&&\\
               &&$\bullet$Noisy Net&&\\
\hline
\multicolumn{5}{c}{\cellcolor{black}{\textcolor{white}{Policy-based method}}} \\
\hline
A3C/A2C & Use actor-critic architecture to estimate&$\bullet$ Multi-step learning&$\circ$ Policy updates exhibit &\cite{65h}\\
        & directly the agent policy. A3C enables &$\bullet$ Actor-critic model&high variance&\cite{65j}\\
        & concurrent learning by allowing multiple &$\bullet$ Advantage function&&\cite{65k}\\
        & learners to operate at the same time      &$\bullet$ Multi-threading&&\\
\hline
UNREAL & Use A3C and multiple unsupervised  &$\bullet$ Unsupervised reward&$\circ$ Policy updates exhibit&\cite{65n}\\
               & reward signals to improve learning efficiency&signals&high variance&\\
               & in complicated environments&&&\\
\hline
DDPG & Concurrently learn a deterministic policy&$\bullet$ Deterministic &$\circ$ Support only continuous&\cite{65j}\\
         & and a Q-function in DQN's fashion&policy gradient&action space&\cite{65k}\\
	 
\hline
TRPO &Limit policy update variance by &$\bullet$ Kullback-Leibler&$\circ$ Computationally expensive&\cite{65j}\\
          &using the conjugate gradient to estimate&divergence&$\circ$ Large batch of rollouts&\cite{65k}\\
          &the natural gradient policy&$\bullet$ Conjugate gradient&$\circ$ Hard to implement&\\
          &TRPO is better than DDPG in terms of&$\bullet$ Natural policy gradient&&\\
          &sample efficiency&&&\\
\hline
ACKTR &Inherit the A2C method&$\bullet$ Kronecker-factored&$\circ$ Still complex&\cite{65j}\\
             &Use Kronecker-Factored approximation to&approximate curvature&&\\
             & reduce computational complexity of TRPO&&&\\
             &ACKTR outperforms TRPO and A2C&&&\\
\hline
ACER & Integrate an experience replay into A3C&$\bullet$ Importance weight&$\circ$ Excessive memory usage&\cite{65j}\\
		  & Introduce a light-weight version of TRPO &truncation \& bias correction&$\circ$ Still complex&\cite{65k}\\
		  & ACER outperforms TRPO and A3C&$\bullet$ Efficient TRPO&&\\
\hline
PPO &Simplify the implementation of TRPO&$\bullet$ Clipped objective&$\circ$ Require network tuning&\cite{65h}\\
        & by using a surrogate objective function &$\bullet$ Adaptive KL penalty&&\cite{65j}\\
        &Achieve the best performance in continuous&coefficient&&\cite{65k}\\
        &control tasks&&&\\
\hline
\hline
\end{tabular}
\end{table*}

\subsection{Single-Agent Method}

The first advent of deep RL, \emph{Deep Q-Network} (DQN) \cite{22, 49}, basically uses a deep neural network to estimate values of state-action pairs via a \emph{Q-value function} (\emph{a.k.a.}, \emph{action-value function} or $Q(s,a)$). Thereafter, a number of variants based on DQN were introduced to improve the original algorithm. Typical extensions can be examined such as \emph{Double DQN} \cite{50}, \emph{Dueling Network} \cite{51}, \emph{Prioritized Experience Replay} \cite{52}, \emph{Recurrent DQN} \cite{53}, \emph{Attention Recurrent DQN} \cite{54}, and an ensemble method named \emph{Rainbow} \cite{55}. These approaches use an experience replay to store historical transitions and retrieve them in batches to train the network. Moreover, a separate network \emph{target network} can be used to mitigate the correlation of the sequential data and prevent the training network from overfitting. 

Instead of estimating the action-value function, we can directly approximate the agent's policy $\pi(s)$. This approach is known as the \emph{policy gradient} or \emph{policy-based} method. \emph{Asynchronous Advantage Actor-Critic} (A3C) \cite{56} is one of the first policy-based deep RL methods to appear in the literature. In particular, A3C includes two networks: an actor network that is used to estimate the agent policy $\pi(s)$ and a critic network that is used to estimate the \emph{state-value function} $V(s)$. Additionally, to stabilize the learning process, A3C uses the \emph{advantage function}, \emph{i.e.}, $A(s,a) = Q(s,a) - V(s)$. There is a synchronous version of A3C, namely A2C \cite{56}, which has the advantage of being simpler but with comparable or better performance. A2C mitigates the risk of multiple learners which might be overlapping when updating the weights of the global networks.

There have been a great number of policy gradient methods since the development of A3C. For instance, \emph{UNsupervised REinforcement and Auxiliary Learning} (UNREAL) \cite{57} uses multiple unsupervised pseudo-reward signals at the same time to improve the learning efficiency in complicated environments. Rather than estimating a stochastic policy, \emph{Deterministic Policy Gradient} \cite{58} (DPG) finds a deterministic policy, which significantly reduces data sampling. Moreover, \emph{Deep Deterministic Policy Gradient} \cite{59} (DDPG) combines DPG with DQN to enable the learning of a deterministic policy in a continuous action space using the actor-critic architecture. The authors in \cite{60} even propose \emph{Multi-agent DDPG} (MADDPG), which employs DDPG in multi-agent environments. To further stabilize the training process, the authors in \cite{61} introduce the \emph{Trust Region Policy Optimization} (TRPO) method, which integrates the \emph{Kullback--Leibler divergence} \cite{62} into the training procedure. However, the implementation of the method is complicated. In 2017, Wu \emph{et al.} \cite{64} proposed \emph{Actor-Critic using Kronecker-Factored Trust Region} (ACKTR), which applies Kronecker-factored approximation curvature into gradient update steps. Additionally, the authors in \cite{65} introduced an efficient off-policy sampling method based on A3C and an experience replay, namely \emph{Actor-Critic with Experience Replay} (ACER). To simplify the implementation of TRPO, ACKTR, and ACER, \emph{Proximal Policy Optimization} (PPO) \cite{63} is introduced by using a clipped ``surrogate" objective function together with stochastic gradient ascent. Finally, some studies combine a policy-based and value-based method such as \cite{65b, 65c, 65d} or an on-policy and off-policy method such as \cite{65e, 65f}. Table \ref{table:2} summarizes key deep RL methods and their reliable implementation repositories. Based on specific application domains, software managers can select a suitable deep RL method to act as a baseline for the target system.

\subsection{Multi-Agent Method}

In multi-agent learning, there are two widely used schemes in the literature: \emph{individual} and \emph{mutual}. In the first case, each agent in the system can be considered as an independent decision maker and other agents as a part of the environment. In this way, any deep RL methods in the previous subsection can be used in multi-agent learning. For instance, Tampuu \emph{et al.} \cite{66} used DQN to create an independent policy for each agent. The authors analyze the behavioral convergence of the involved agents with respect to cooperation and competition. Similarly, Leibo \emph{et al.} \cite{67} introduced a sequential social dilemma, which basically uses DQN to analyze the agent's strategy in Markov games such as Prisoner's Dilemma, Fruit Gathering, and Wolfpack. However, the approach limits the number of agents because the computational complexity increases with the number of policies. To overcome this obstacle, Nguyen \emph{et al.} developed a behavioral control system \cite{48}  in which homogeneous agents can share the same policy. As a result, the method is robust and scalable. Another problem in multi-agent learning is the use of an experience replay, which amplifies the non-stationary problem that occurs due to asynchronous data sampling of different agents \cite{68, 69}. A lenient approach \cite{70} can subdue the problem by mapping transitions into decaying temperature values, which basically controls the magnitude of updating different policies.

In mutual scheme, agents can ``speak" with each other via a settled communication channel. Moreover, agents are often trained in a centralized manner but eventually operate in a decentralized fashion when deployed \cite{71}. In other words, a multi-agent RL problem can be divided into two sub-problems: a goal-directed problem and a communication problem. Specifically, Foerster \emph{et al.} \cite{72} introduced two communication schemes based on the centralized-decentralized rationale: \emph{Reinforced Inter-Agent Learning} (RIAL) and \emph{Differentiable Inter-Agent Learning} (DIAL). While RIAL reinforces agents' learning by sharing parameters, DIAL allows \emph{inter-communication} between agents via a shared medium. Both methods, however, operate with a discrete number of communication actions. As opposed to RIAL and DIAL, the authors in \cite{73} introduce a novel network architecture, namely \emph{Communication Neural Net} (CommNet), which enables communication by using a continuous vector. As a result, the agents are trained to learn to communicate by backpropagation. However, CommNet limits the number of agents due to the increase of computational complexity. To make it scalable, Gupta \emph{et al.} \cite{74} introduced a parameter sharing method that can handle a large number of agents. However, the method only works with homogeneous systems. Finally, Nguyen \emph{et al.} \cite{48} extended the Gupta's study to heterogeneous systems by designing a behavioral control system. For rigorous study, a complete survey on multi-agent RL can be found in \cite{75,76,77}.

In summary, it is critical to address the following factors in multi-agent learning because they have a great impact on the target software architecture:

\begin{itemize}
\item It is preferable to employ the centralized-decentralized rationale in a multi-agent RL-based system because the training process is time-consuming and computationally expensive. A working system may require hundreds to thousands of training sessions by searching through the hyper-parameter space to find the optimal solution.

\item The communication between agents can be \emph{realistic} or \emph{imaginary}. In realistic communication, agents ``speak" with each other using an established communication protocol. However, there is no actual channel in imaginary communication. The agents are trained to collaborate using a specialized network architecture. For instance, OpenAI \cite{78} proposes an actor-critic architecture where the critic is augmented with other agents' policy information. As a result, the two methods can differentiate how to design an RL-based system.

\item A partially observable environment has a great impact on designing a multi-agent system because each agent has its own unique perspective of the environment. Therefore, it is important to first carefully examine the environment and application type to avoid any malfunction in the design.

\end{itemize}

\subsection{RL Challenges}
In this subsection, we briefly review major challenges while designing a deep RL-based system and corresponding solutions. To remain concise, the proposed framework is not concerned with these techniques but it is straightforward to extend the architecture to support these rectifications.

\emph{Catastrophic forgetting} is a problem that occurs in \emph{continual learning} and \emph{multi-task learning}, \emph{i.e.}, an another task is trained after learning the first task by using the same neural network. In this case, the neural network gradually forgets the knowledge of the first task to adopt the new one. One solution is to use regularization \cite{79, 80} or a dense neural network \cite{81, 82}. However, these approaches are only feasible with a limited number of tasks. Recent studies introduce more scalable approaches such as \emph{Elastic Weight Consolidation} (EWC) \cite{86} or \emph{PathNet} \cite{86b}. While EWC finds a configuration of the network that yields the best performance in different tasks, PathNet uses a ``super" neural network to fulfill the knowledge of different tasks in different paths. 

\emph{Policy distillation} \cite{83} or \emph{transfer learning} \cite{84, 85} can be used to train an agent to learn individual tasks and collectively transfer the knowledge to a single network. Transfer learning is often used when the actual experiment is expensive and intractable. In this case, the network is trained with simulations and later is deployed into the target experiment. However, a \emph{negative transfer} may occur when the performance of the learner is lower than the trainer. The authors in \cite{87} introduced \emph{Hierarchical Prioritized Experience Replay} that uses high-level features of the task and selects important data from the experience replay to mitigate the negative transfer. One recent study \cite{88} aligned the mutual learning to achieve a comparable performance between the actual experiment and simulations. 

Another obstacle in RL is dealing with \emph{long-horizon} environments with \emph{sparse rewards}. In such tasks, the agent hardly receives any reward and easily gets stuck in local minimum solutions. One straightforward solution is to use \emph{reward shaping} \cite{89} that continuously instructs the agent to achieve the objective. The problem can also be divided into a hierarchical tree of sub-problems where the parent problem has a higher abstraction than the child problem (\emph{Hierarchical RL}) \cite{90}. To encourage self-exploration, the authors in \cite{91} introduced intrinsic reward signals to reinforce the agent to make a generic decision. State-of-the-art methods of \emph{intrinsic motivation} can be found in \cite{92, 93, 94}. Finally, Andrychowicz \emph{et al.} \cite{95} propose \emph{Hindsight Experience Replay} that implicitly simulates \emph{curriculum learning} \cite{96} by creating imagined trajectories in the experience replay with positive rewards. In this way, the agent can learn from failures and automatically generalize a solution in success cases.

Finally, a variety of RL-related techniques are proposed to make RL feasible in large-scale applications. One approach is to augment the neural network with a ``memory" to enhance sample efficiency in complicated environments \cite{97, 98}. Additionally, to enforce scalability, many distributed methods have been proposed such as \emph{Distributed Experience Replay} \cite{99}, deep RL acceleration \cite{100}, and distributed deep RL \cite{101}. Finally, \emph{imitation learning} can be used together with \emph{inverse RL} to speed up training by directly learning from expert demonstrations and extracting the expert's cost function \cite{102}.

\subsection{Deep RL Framework}

In this subsection, we discuss the latest deep RL frameworks in the literature. We select the libraries based on different factors including Python-based implementation, clear documentation, reliability, and active community. Based on our analysis, software managers can select a suitable framework depending on project requirements.

\begin{figure}[!t]
\centering
\includegraphics[width=0.85\linewidth]{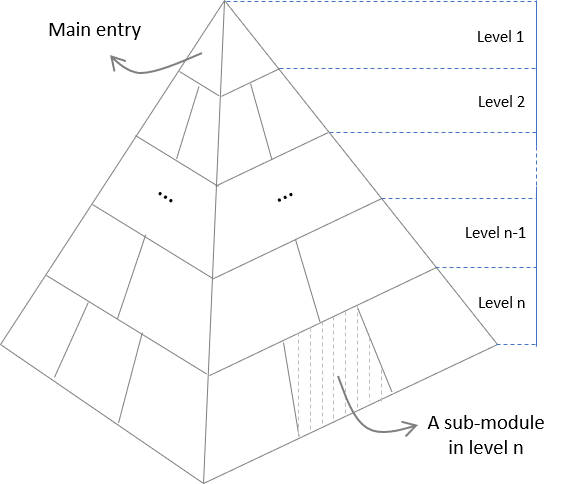}
\caption{A ``pyramid" software architecture.}
\label{fig:4} 
\end{figure}

\begin{itemize}

\item \emph{Chainer} -- Chainer \cite{65k} is a powerful and flexible framework for neural networks. The framework is currently supported by IBM, Intel, Microsoft, and Nvidia. It provides an easy way to manipulate neural networks such as by creating a customized network, visualizing a computational graph, and supporting a debug mode. It also implements a variety of deep RL methods. However, the Chainer's architecture is complicated and requires a great effort to develop a new deep RL method. The number of integrated environments is also limited, \emph{e.g.}, Atari \cite{103}, OpenAI Gym \cite{104}, and Mujoco \cite{105}.

\item \emph{Keras-RL} -- Keras-RL \cite{106} is a friendly deep RL library, which is recommended for deep RL beginners. However, the library provides a limited number of deep RL methods and environments.  

\item \emph{TensorForce} -- TensorForce \cite{107} is an ambitious project that targets both industrial applications and academic research. The library has the best modular architecture we have reviewed so far. Therefore, it is convenient to use the framework to integrate customized environments, modify network configurations, and tweak deep RL algorithms. However, the framework has a deep software stack (``pyramid" model) that includes many abstraction layers, as shown in Fig.~\ref{fig:4}. This hinders novice readers from prototyping a new deep RL method.

\item \emph{OpenAI Baselines} -- OpenAI Baselines \cite{65j} is a high-quality framework for contemporary deep RL. In contrast to TensorForce, the library is suitable for researchers who want to reproduce original results. However, OpenAI Baselines is unstructured and incohesive.

\item \emph{RLLib} -- RLLib \cite{65h} is a well-designed deep RL framework that provides a means to deploy deep RL in distributed systems. Therefore, the usage of RLLib is not friendly for RL beginners. 

\item \emph{RLLab} -- RLLab \cite{108} provides a diversity of deep RL methods including TRPO, DDPG, Cross-Entropy Method, and Evolutionary Strategy. The library is friendly to use but not straightforward for modifications.

\end{itemize}

In summary, most deep RL frameworks focus on the performance of deep RL methods. As a result, those frameworks limits code legibility, which basically restricts RL beginners from readability and modifications. In this paper, we propose a comprehensive framework that has the following properties:

\begin{itemize}

\item Allow new users to prototype a deep RL method in a short period of time by following a modular design. As opposed to TensorForce, we limit the number of abstraction layers and avoid the pyramid structure.

\item The framework is friendly with a simplified user interface. We provide an API based on three key concepts: policy network, network configuration, and learner.

\item Enforce scalability and generalization while supporting multiple agents, multiple objectives, and human-machine interactions.

\item Finally, we introduce a concept of unification and transparency by creating plugins. Plugins are gateways that extract learners from other libraries and plug them into our proposed framework. In this way, users can interact with different frameworks using the same interface.

\end{itemize}

\section{Software Architecture}
\label{sec:3}

In this section, we examine core components towards designing a comprehensive deep RL framework, which basically employs generality, flexibility, and interoperability. We aim to support a broad range of RL-related applications that involve multiple agents, multiple objectives, and human-agent interaction. We use the following pseudocode to describe a function signature:

\begin{center}
\fbox{%
	\parbox{0.95\linewidth}{
		\begin{multline*}
		\bullet\textbf{function\_name}([param1, param2,...])\rightarrow \\ 
		[return1,return2,...]\text{ or }\{return1, return2, ...\} \text{ or } A
		\end{multline*}
	}
}
\end{center}

\noindent
where $\rightarrow$ denotes a return operation, $A$ is a scalar value, $[...]$ denotes an array, and $\{...\}$ denotes a list of possible values of a single variable.

\subsection{Environment}
\label{sec:3.1}

\begin{figure}[!t]
\centering
\includegraphics[width=0.95\linewidth]{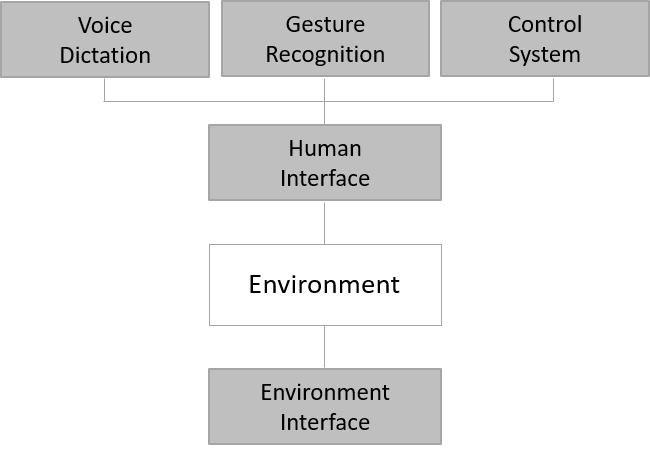}
\caption{A conceptual model of the environment with a human interface.}
\label{fig:3} 
\end{figure}

First, we create a unique interface for the \emph{environment} to establish a communication channel between the framework and agents. However, to reduce complexity, we put any human-related communication into the environment. As a result, human interactions can be seen as a part of the environment and are hidden from the framework, \emph{i.e.}, the environment provides two interfaces: one for the framework and one for human, as shown in Fig.~\ref{fig:3}. While the framework interface is often in programming level (functions), the human interface has a higher abstraction mostly in human understanding forms such as voice dictation, gesture recognition, or control system.

Essentially, the environment's framework interface should provide the following functions:

\begin{itemize}
\item \textbf{clone}(): the environment can duplicate itself. The function is useful when an RL algorithm requires multiple learners at the same time (\emph{e.g.} A3C).

\item \textbf{reset}(): reset the environment to the initial state. The function must be called after or before an episode.

\item \textbf{step}($[a_1,a_2,...,a_N]$) $\rightarrow$ [$r_1,r_2,...,r_M$]: executes $N$ specified actions of $N$ agents in the environment. The function returns $M$ rewards, each of which represents an objective function.

\item \textbf{get\_state}() $\rightarrow$ [$s_1,s_2,...,s_N$]: retrieves the current states of the environment. If the environment is a partially observable MDP, the function returns $N$ states, which each presents the current state of an agent. However, if the environment is a fully observable MDP, we have $s_1 = s_2 = ... = s_N = s$.

\item \textbf{is\_terminal}() $\rightarrow$ \{True, False\}: checks if the episode is terminated.

\item \textbf{get\_number\_of\_objectives}() $\rightarrow$ $M$: is a helper function that indicates the number of objectives in the environment.

\item \textbf{get\_number\_of\_agents}() $\rightarrow$ $N$: is a helper function that indicates the number of agents in the environment.

\end{itemize}

Finally, it is important to consider the following questions while designing an environment component as they have a great impact on subsequent design stages:

\begin{itemize}
\item \emph{Is it a simulator or a wrapper?} In the case of a wrapper, the environment is already developed and configured. Our duty is to develop a wrapper interface that can compatibly interact with the framework. In contrast to the wrapper, developing a simulator is complicated and requires expert knowledge. In real-time applications, we may first develop a simulator in C/C++ (for better performance) and then create a Python wrapper interface (for easier integration). In this case, we need to develop both a simulator and a wrapper.  

\item \emph{Is it stochastic or deterministic?} Basically, a stochastic environment is more challenging to implement than a deterministic one. There are potential factors that are likely to contribute to the randomness of the environment. For example, a company intends to open a bike rental service. \emph{N} bikes are equally distributed into \emph{M} potential places. However, at a specific time, place \emph{A} still has a plenty of bikes because there is no customer. As a result, bikes in place \emph{A} are delivered to other places where the demand is higher. The company seeks development of an algorithm that can balance the number of bikes in each place over time. In this example, the bike rental service is a stochastic environment. We can start building a simple stochastic model based on Poisson distribution to represent the rental demand in each place. We end up with a complicated model based on a set of observable factors such as rush hour, weather, weekend, festival, etc. Depending on the stochasticity of the model, we can decide to use a model-based or model-free RL method.

\item \emph{Is it complete or incomplete?} A complete environment at any time provides sufficient information to construct a branch of possible moves in the future (\emph{e.g.} Chess or Go). The completeness can help to decide an effective RL method later. For instance, a complete environment can be solved with a careful planning rather than a trial-and-error approach.

\item \emph{Is it fully observable or partially observable?} The observability of environment is essential when designing a deep neural network. Partially observable environments might require recurrent layers or an attention mechanism to enhance the network's capacity during the training. A self-driving scenario is partially observable while a board game is fully observable.

\item \emph{Is it continuous or discrete?} As described in Table \ref{table:1}, this factor is important to determine the type of methods used such as policy-based or value-based methods or the network configuration such as actor-critic architectures.

\item \emph{How many objectives does it have?} Real-world applications often have multiple objectives. If the importance weights between objectives can be identified in the beginning, it is reasonable to use a single-policy RL method. Alternatively, a multi-policy RL method can prioritize the importance of an objective in real time.

\item \emph{How many agents does it have?} A multi-agent RL-based system is much more complicated than a single-agent counterpart. Therefore, it is essential to analyze the following factors of a multi-agent system before delving into the design: the number of agents, the type of agents, communication abilities, cooperation strategies, and competitive potentials. 

\end{itemize}

\subsection{Network}

The \emph{neural network} is also a key module of our proposed framework, which includes a network configuration and a policy network, as illustrated in Fig. \ref{fig:5}. A network configuration defines the deep neural network architecture (\emph{e.g.} CNN or LSTM), loss functions (\emph{e.g.} Mean Square Error or Cross Entropy Loss), and optimization methods (\emph{e.g.} Adam or SGD). Depending on the project's requirements, a configuration can be divided to different abstraction layers, where the lower abstraction layer is used as a mapping layer for the higher abstraction layer. In the lowest abstraction level (programming language), a configuration is implemented by a deep learning library, such as Pytorch \cite{107b} (with dynamic graph) or TensorFlow (with static graph). The next layer is to use a scripting language, such as \emph{xml} or \emph{json}, to describe the network configuration. This level is useful because it provides a faster and easier way to configure a network setting. For those who do not have much knowledge of implementation details such as system analysts, a graphical user interface can assist them. However, there is a trade-off here: the higher abstraction layer achieves better usability and productivity but has a longer development cycle.

A policy network is a composite component that includes a number of network configurations. However, the dependency between a policy network and a configuration can be weak, \emph{i.e.}, an \emph{aggregation} relationship. The policy network's objective is twofold. It provides a high-level interface that maintains connectivity with other modules in the framework, and it initializes the network, saves the network's parameters into checkpoints, and restores the network's parameters from checkpoints. Finally, the neural network interface should provide the following functions:

\begin{figure}[!t]
\centering
\includegraphics[width=0.93\linewidth]{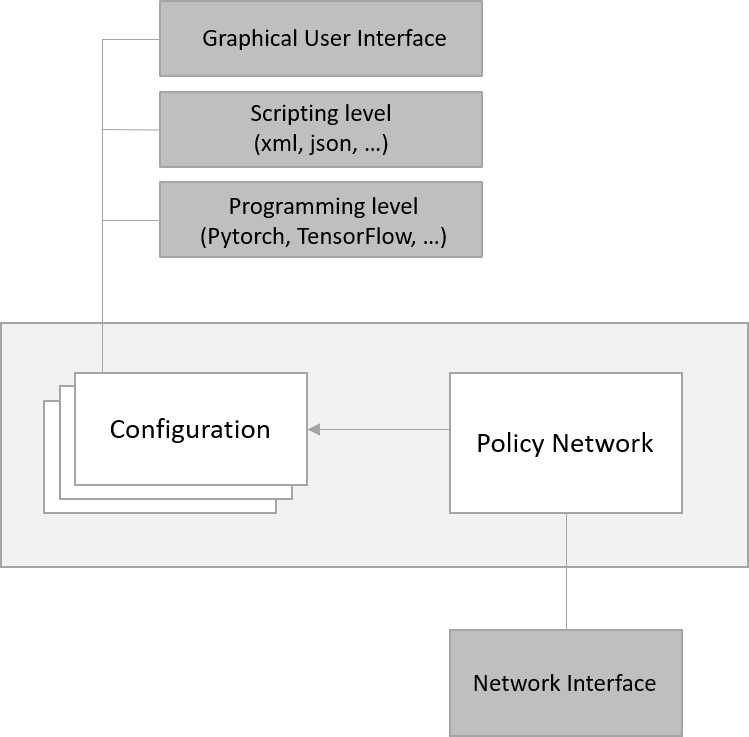}
\caption{A neural network module includes a network configuration and a policy network.}
\label{fig:5} 
\end{figure}

\begin{itemize}

\item \textbf{create\_network}() $\rightarrow$ [$\theta_1, \theta_2, ..., \theta_K$]: instantiates a deep neural network by using a set of network configurations. The function returns the network's parameters (references) $\theta_1, \theta_2, ..., \theta_K$.

\item \textbf{save\_model}(): saves the current network's parameters into a checkpoint file.

\item \textbf{load\_model}([chk]): restores the current network's parameters from a specified checkpoint file \emph{chk}.

\item \textbf{predict}([$s_1,s_2,...,s_N$]) $\rightarrow$ [$a_1, a_2, ..., a_N$]: given the current states of $N$ agents $s_1,s_2,..._,s_N$, the function uses the network to predict the next $N$ actions $a_1, a_2, ..., a_N$.

\item \textbf{train\_network}([data\_dict]): trains the network by using the given data dictionary. The data dictionary often includes the current states, current actions, next states, terminal flags, and miscellaneous information (global time step or objective weights) of $N$ agents.

\end{itemize}

\subsection{Learner}

\begin{figure}[!t]
\centering
\includegraphics[width=0.9\linewidth]{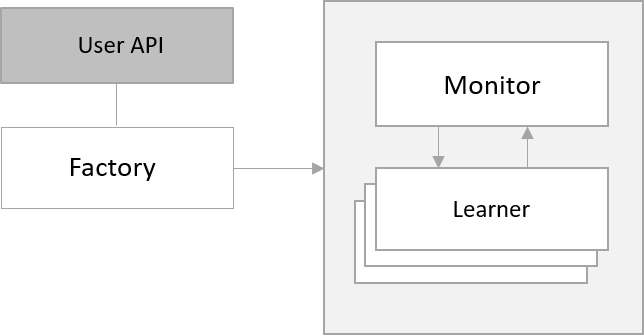}
\caption{A high-level design of a learner module.}
\label{fig:6} 
\end{figure}

The last key module of our proposed framework is a \emph{learner}, as shown in Fig. \ref{fig:6}. While the environment module and the network module create the application's shell, the learner plays as an engine that allows the system to operate properly. The three modules jointly create the backbone of the system. In particular, the learner uses the environment module to generate episodes. It manages the experience replay memory and defines the RL implementation details, such as multi-step learning, multi-threading, or reward shaping. The learner is often created together with a monitor. The monitor is used to manage multiple learners (if multi-threading is used) and collect any data from the learners during training, such as performance information for debugging purposes and post-evaluation reports. Finally, the learner collects necessary data, packs it into a dictionary, and sends it to the network module for training.

Additionally, a \emph{factory} pattern \cite{107c} can be used to hide the operating details between the monitor and the learners. As a result, the factory component promotes higher abstraction and usability through a simplified user API as below:

\begin{itemize}

\item \textbf{create\_learner}([monitor\_dict, learner\_dict]) $\rightarrow$ obj: The factory creates a learner by using the monitor's data dictionary (batch size, the number of epochs, report frequency, etc) and the learner's data dictionary (the number of threads, epsilon values, reward clipping thresholds, etc.).

\item \textbf{train()}: trains the generated learner.

\item \textbf{evaluate()}: evaluates the generated learner.

\end{itemize}

\subsection{Plugin}

\begin{figure}[!t]
\centering
\includegraphics[width=0.98\linewidth]{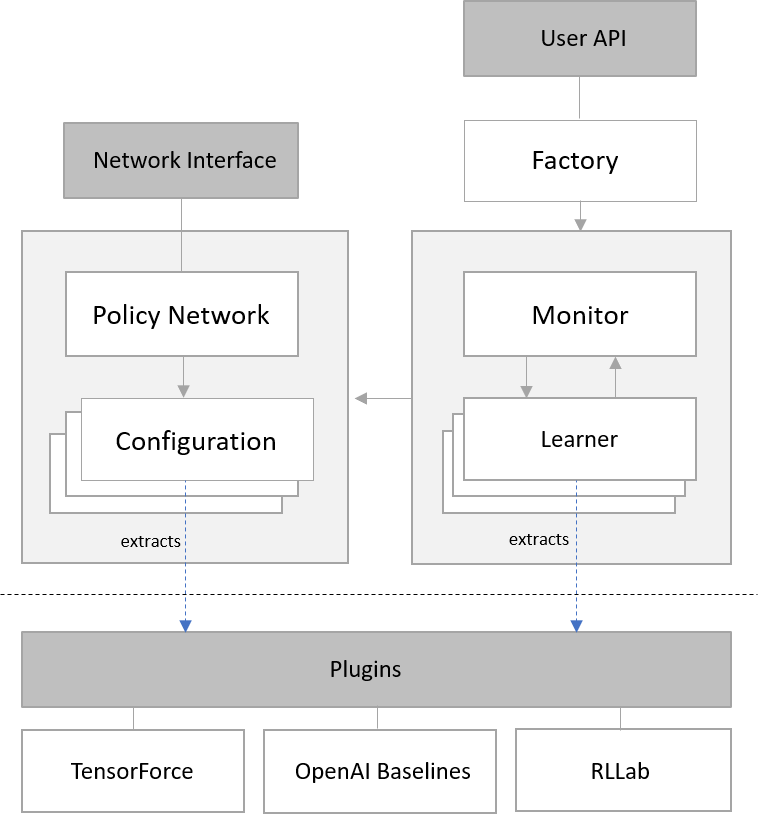}
\caption{A unification of different RL libraries by using plugins.}
\label{fig:7} 
\end{figure}

There have been a great number of RL methods in the literature. Therefore, it is impractical to implement all of them. However, we can reuse the implementation from existing libraries such as TensorForce, OpenAI Baselines, or RLLab. To enforce flexibility and interoperability, we introduce a concept of unification by using plugins. A plugin is a piece of program that extracts learners or network configurations from third party libraries and plugs them into our framework. As a result, the integrated framework provides a unique user API but supports a variety of RL methods. In this way, users do not need to learn different libraries. The concept of unification is described in Fig. \ref{fig:7}.

A plugin can also act as a conversion program that converts the environment's interface of this library into the environment's interface of other libraries. As a result, the proposed framework can work with any environments in third party libraries and vice versa. Therefore, a plugin should include the following functions:

\begin{itemize}

\item \textbf{convert\_environment}([source\_env]) $\rightarrow$ target\_env: converts the environment's interface from the source library to the environment's interface defined in the target library.

\item \textbf{extract\_learner}([param\_dict])  $\rightarrow$ learner: extracts the learner from the target library.

\item \textbf{extract\_configuration}([param\_dict]) $\rightarrow$ config: extracts the network configuration from the target library.

\end{itemize}

\subsection{Overall Structure}

\begin{figure*}[!t]
\centering
\includegraphics[width=1.\linewidth]{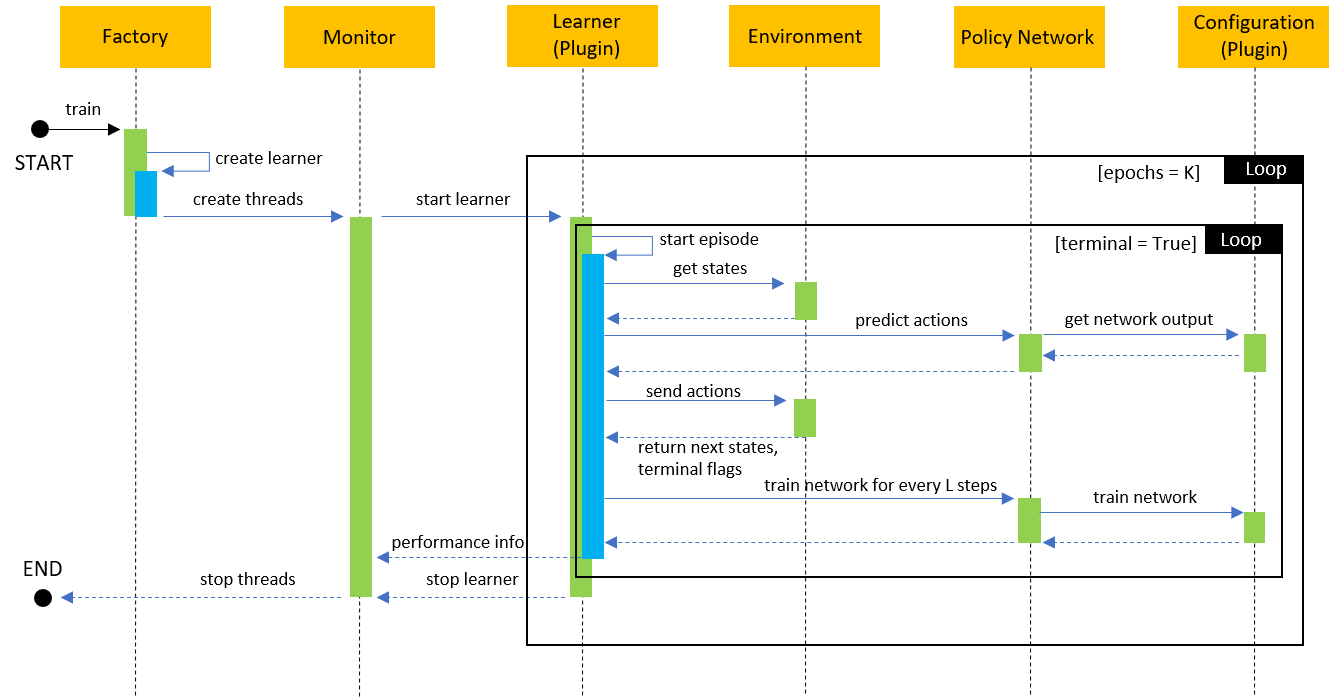}
\caption{A UML sequential diagram of the training process.}
\label{fig:8} 
\end{figure*}

\begin{figure*}[!t]
\centering
\includegraphics[width=1.\linewidth]{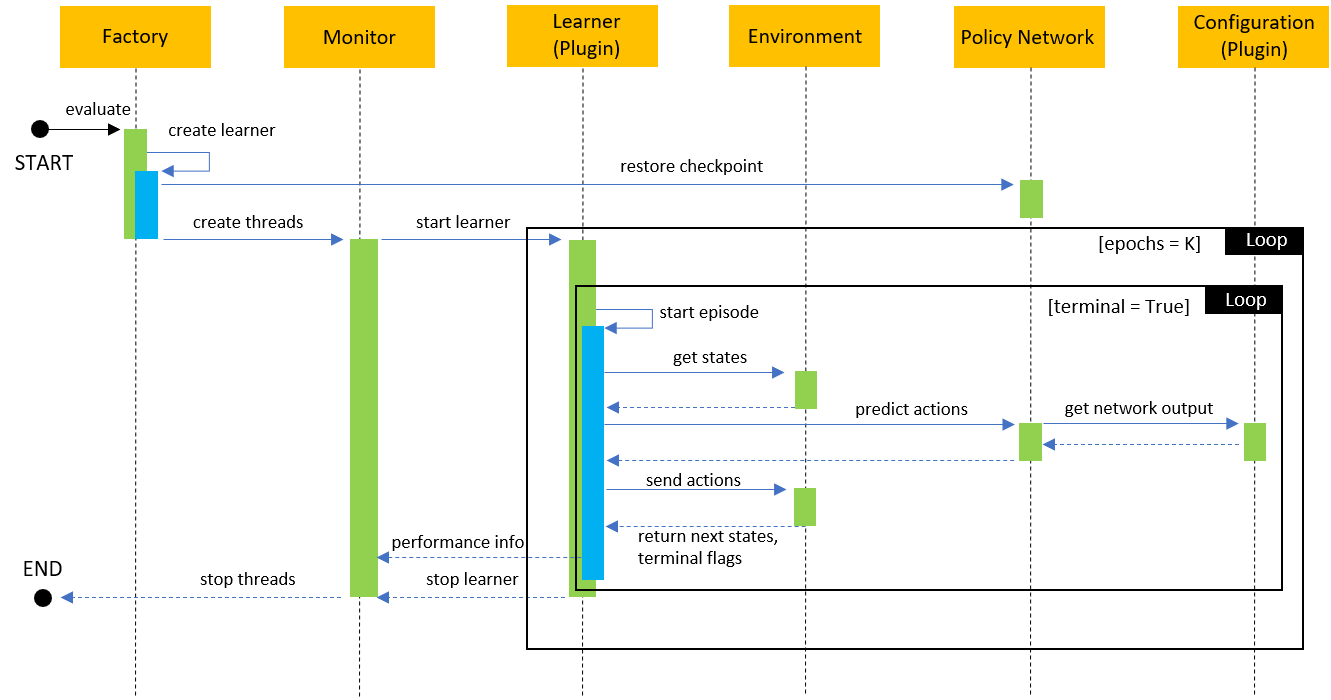}
\caption{A UML sequential diagram of the evaluation process.}
\label{fig:8b} 
\end{figure*}

\begin{table*}[!t]
\renewcommand{\arraystretch}{1.3}
\caption{Demonstration codes of different use cases \cite{109}.}
\label{table:3}
\centering
\begin{tabular}{ccc}
\hline
\hline
\textbf{Use case} & \textbf{Description} & \textbf{Source Code}\\
\hline
1. How to inherit an existing learner? & Develop a Monte-Carlo learner that inherits the & fruit/learners/mc.py\\
&existing Q-Learning learner&\\
\hline
2. Develop a new environment & Create a Grid World \cite{1} environment that follows& fruit/envs/games/grid\_world/\\
& the framework's interface&\\
\hline
3. Develop a multi-agent environment & Create a Tank Battle \cite{48} game in which& fruit/envs/games/tank\_battle/\\
with human-agent interaction &humans and AI agents can play together&\\
\hline
4. Multi-objective environment and & Use a multi-objective learner (MO Q-Learning)& fruit/samples/basic/multi\_objectives\_test.py\\
multi-objective RL&to train an agent to play Mountain Car \cite{41}&\\
\hline
5. Multi-agent learner with & Create a multi-agent RL method based on A3C \cite{74}& fruit/samples/tutorials/chapter\_6.py \\
human-agent interaction &and apply it to Tank Battle&\\
\hline
6. How to use a plugin? &Develop a TensorForce plugin, extract the PPO learner,&fruit/plugins/quick\_start.py \\
&and train an agent to play Cart Pole \cite{1}& \\
\hline
\hline
\end{tabular}
\end{table*}

Putting everything together, we have a sequential diagram of the training process, as described in Fig. \ref{fig:7}. The workflow breaks down the training process into smaller procedures. Firstly, the factory instantiates a specified learner (or a plugin) and sends its reference to the monitor. The monitor clones the learner into multiple learner threads. Each learner thread is run until the number of epochs exceeds a predefined value, $K$. The second loop within the learner thread is used to generate episodes. In each episode, a learner thread perceives the current states of the environment and predicts next actions using the policy network and configuration network. The next actions are applied to the environment. The environment returns next states and a terminal flag. Finally, the policy network is trained for every $L$-step. There are minor changes in the evaluation process, as shown in Fig.~\ref{fig:8b}. First, the policy network's parameters are restored from a specified checkpoint file while initializing the learner. Second, all training procedure calls are discarded while generating episodes.

\begin{figure}[!t]
\centering
\includegraphics[width=0.65\linewidth]{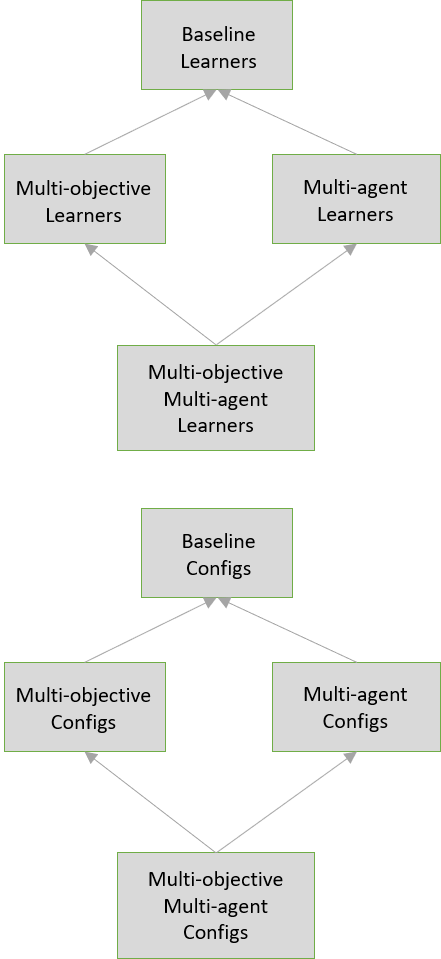}
\caption{An inheritance relationship between learners and configurations.}
\label{fig:9} 
\end{figure}

To enhance usability and reduce redundancy, it is advisable to implement the framework in \emph{Object-Oriented Programming} (OOP). In this way, a new learner (configuration) can be easily developed by inheriting existing learners (configurations) in the framework, as shown in Fig. \ref{fig:9}.

\section{Conclusions}
\label{sec:4}
It is promising to see many successful applications of deep RL methods in recent years. This paper has briefly reviewed recent advances in the RL literature with respect to multi-agent learning, multi-objective learning, and human-machine interactions. We also examine different deep RL libraries and analyze their limitations. Finally, we propose a novel design of a deep RL framework that exhibits great potential in terms of usability, flexibility, and interoperability. We highlight any considerable notices during the design so that software managers can avoid possible mistakes while designing an RL-based application.

The proposed framework can be considered as a \emph{template} to design a real-world RL-based application. Because the framework is developed in OOP, it is beneficial to utilize OOP principles, such as inheritance, polymorphism, and encapsulation to expedite the whole development process. We advisedly created a ``soft'' software layer stack, where the number of modules is minimal while maintaining a certain level of cohesion. As a result, 
the learning curve is not steep. By providing a simplified API, the framework is suitable for novice readers who are new to deep RL, especially software engineers. Finally, the framework acts as a bridge to connect different RL communities around the world.

Our ultimate goal is to build an educational software platform for deep RL. The next development milestone includes three steps:

\begin{itemize}

\item Implement a variety of plugins for the proposed framework.

\item Develop a GUI application that can configure the neural network, modify the learner, and visualize the RL workflow.

\item Complete documentation including tutorials and sample codes.
\end{itemize}

\section*{Appendix}

To keep the paper brief, we provide documentation of the proposed framework as online materials \cite{108}. These include an installation guide, code samples, benchmark scores, tutorials, an API reference guide, a class diagram, and a package diagram. Table \ref{table:3} lists demonstration codes of different use cases (codebase \cite{109}).

\section*{Acknowledgement}

The authors wish to thank our colleagues in the Institute for Intelligent Systems Research and Innovation for their comments and helpful discussions. We truly appreciate Nguyen Chau, a principle IT product manager at Atlassian, who shared his expertise in the field to eradicate any misunderstandings in this paper. We also thank Dr. Thanh Nguyen, University of Chicago, for being an active adviser in the design process. Finally, we send our gratefulness to the RL community who provided crucial feedback during the project's beta testing.





\begin{thebibliography}{1}


\bibitem{1}
R.~S.~Sutton and G.~B.~Andrew, \emph{Introduction to Reinforcement Learning}. Cambridge: MIT press, 1998.

\bibitem{2}
N.~D.~Nguyen, T.~Nguyen, and S.~Nahavandi, ``System design perspective for human-level agents using deep reinforcement learning: A survey," \emph{IEEE Access}, vol.~5, pp.~27091--27102, 2017.

\bibitem{3}
H.~Mao, M.~Alizadeh, I.~Menache, and S.~Kandula, ``Resource management with deep reinforcement learning," In \emph{Proceedings of the 15th ACM Workshop on Hot Topics in Networks}, pp.~50--56, 2016.

\bibitem{4}
T.~T.~Nguyen, and V.~J.~Reddi, ``Deep reinforcement learning for cyber security," \emph{arXiv preprint arXiv:1906.05799}, 2019.

\bibitem{5}
D.~Fox, W.~Burgard, H.~Kruppa, and S.~Thrun, ``A probabilistic approach to collaborative multi-robot localization," \emph{Autonomous Robots}, vol.~8, no.~3, pp.~325--344, 2000.

\bibitem{6}
M.~Riedmiller, T.~ Gabel, R.~Hafner, and S.~Lange, ``Reinforcement learning for robot soccer," \emph{Autonomous Robots}, vol.~27, no.~1, pp.~55--73, 2009.

\bibitem{6b}
K.~M\"ulling, J.~Kober, O.~Kroemer, and J.~Peters, ``Learning to select and generalize striking movements in robot table tennis," \emph{International Journal of Robotics Research}, vol.~32, no.~3, pp.~263--279, 2013.

\bibitem{7}
T.~G.~Thuruthel et al., ``Model-based reinforcement learning for closed-loop dynamic control of soft robotic manipulators," \emph{IEEE Transactions on Robotics}, vol.~35, pp.124--134, 2018.

\bibitem{7b}
T.~Nguyen, N.~D.~Nguyen, F.~Bello, and S.~Nahavandi, ``A new tensioning method using deep reinforcement learning for surgical pattern cutting," In \emph{2019 IEEE International Conference on Industrial Technology (ICIT)}, pp. 1339-1344. IEEE, 2019.

\bibitem{7c}
N.~D.~Nguyen, T.~Nguyen, S.~Nahavandi, A.~Bhatti, and G.~Guest, ``Manipulating soft tissues by deep reinforcement learning for autonomous robotic surgery," In \emph{2019 IEEE International Systems Conference (SysCon)}, pp. 1-7. IEEE, 2019.

\bibitem{8}
R.~H.~Crites and A.~G.~Barto, ``Elevator group control using multiple reinforcement learning agents," \emph{Machine Learning}, vol.~33, no.~2--3, pp.~235--262, 1998.

\bibitem{9}
I.~Arel, C.~Liu, T.~Urbanik, and A.~G.~Kohls, ``Reinforcement learning-based multi-agent system for network traffic signal control," \emph{IET Intelligent Transport Systems}, vol~4, no.~2, pp.~128--135, 2010.

\bibitem{10}
G.~Zheng, F.~Zhang, Z.~Zheng, Y.~Xiang, N.~J.~Yuan, X.~Xie, and Z.~Li, ``DRN: A deep reinforcement learning framework for news recommendation," In \emph{Proceedings of the 2018 World Wide Web Conference}, pp.~167--176, 2018.

\bibitem{11}
J.~Jin, C.~Song, H.~Li, K.~Gai, J.~Wang, and W.~Zhang, ``Real-time bidding with multi-agent reinforcement learning in display advertising.," In \emph{Proceedings of the 27th ACM International Conference on Information and Knowledge Management}, pp.~2193--2201, 2018.

\bibitem{12}
M.~Campbell, A.~J.~Hoane, and F.~H.~Hsu, ``Deep blue," \emph{Artificial Intelligence}, vol.~134, no.~1--2, pp.~57--83, 2002.

\bibitem{13}
G.~Tesauro and G.~R.~Galperin, ``On-line policy improvement using monte-carlo search," in \emph{Advances in Neural Information Processing Systems}, pp.~1068--1074, 1997.

\bibitem{14}
G.~Tesauro, ``Temporal difference learning and td-gammon," \emph{Communication}, vol.~38, no.~3, pp.~58--68, 1995.

\bibitem{14b}
T.~Nguyen, N.~D.~Nguyen, and S.~Nahavandi, ``Multi-agent deep reinforcement learning with human strategies," In \emph{2019 IEEE International Conference on Industrial Technology (ICIT)}, pp. 1357-1362. IEEE, 2019.

\bibitem{14c}
N.~D.~Nguyen, S.~Nahavandi, and T.~Nguyen. ``A human mixed strategy approach to deep reinforcement learning," In \emph{2018 IEEE International Conference on Systems, Man, and Cybernetics (SMC)}, pp. 4023-4028. IEEE, 2018.

\bibitem{15}
R.~Bellman, \emph{Dynamic Programming}. Princeton: Princeton University Press, 2010.

\bibitem{16}
M.~Fowler and K.~Scott, \emph{UML Distilled: A Brief Guide to the Standard Object Modeling Language}. Addison-Wesley Professional, 2004.

\bibitem{17}
T.~J.~Ross, \emph{Fuzzy Logic with Engineering Applications}. John Wiley \& Sons, 2005.

\bibitem{18}
M.~Hausknecht, J.~Lehman, R.~Miikkulainen, and P.~Stone, ``A neuroevolution approach to general atari game playing," IEEE Transactions on Computational Intelligence and AI in Games, vol.~6, no.~4, pp.~355--366, 2014.

\bibitem{18b}
T.~Salimans, J.~Ho, X.~Chen, A.~Sidor, and I.~Sutskever, ``Evolution strategies as a scalable alternative to reinforcement learning, \emph{arXiv preprint arXiv:1703.03864}, 2017.

\bibitem{19}
D.~P.~Bertsekas, \emph{Dynamic Programming and Optimal Control}. Belmont, MA: Athena scientific, 1995.

\bibitem{20}
J.~Duchi and Y.~Singer, ``Efficient online and batch learning using forward backward splitting," \emph{Journal of Machine Learning Research}, vol.~10, pp.~2899--2934, 2009.

\bibitem{21}
S.~Adam, L.~Busoniu, and R.~Babuska, ``Experience replay for real-time reinforcement learning control, \emph{IEEE Transactions on Systems, Man, and Cybernetics, Part C (Applications and Reviews)}, vol.~42, no.~2, pp.~201--212, 2011.

\bibitem{22}
V.~Mnih et al., ``Human-level control through deep reinforcement learning," \emph{Nature}, 2015.

\bibitem{23}
A.~Krizhevsky, S.~Ilya, and E.~H.~Geoffrey, ``Imagenet classification with deep convolutional neural networks." In \emph{Advances in Neural Information Processing Systems}, 2012.

\bibitem{24}
D.~Silver et al., ``Mastering the game of Go with deep neural networks and tree search," \emph{Nature}, vol.~529, no~7587, 2016.

\bibitem{25}
C.~B.~Browne, E.~Powley, D.~Whitehouse, S.~M.~Lucas, P.~I.~Cowling, P.~Rohlfshagen, S.~Tavener, D.~Perez, S.~Samothrakis, and S.~Colton, ``A survey of monte carlo tree search methods," \emph{IEEE Transactions on Computational Intelligence and AI in games}, vol.~4, no.~1, pp.~1--43, 2011.

\bibitem{26}
G.~Tesauro, ``TD-Gammon, a self-teaching backgammon program, achieves master-level play, \emph{Neural Computation}, vol.~6, no.~2, pp.~215--219, 1992.

\bibitem{27}
A.~E.~Sallab, M.~Abdou, E.~Perot, and S.~Yogamani, ``Deep reinforcement learning framework for autonomous driving," \emph{Electronic Imaging}, vol.~19, pp.~70--76, 2017.

\bibitem{27b}
S.~S.-Shwartz, S.~Shaked, and S.~Amnon, ``Safe, multi-agent, reinforcement learning for autonomous driving," \emph{arXiv preprint arXiv:1610.03295}, 2016.

\bibitem{28}
A.~Y.~Ng, A.~Coates, M.~Diel, V.~Ganapathi, J.~Schulte, B.~Tse, E.~Berger, and E.~Liang, ``Autonomous inverted helicopter flight via reinforcement learning," \emph{Experimental Robotics IX}, pp.~363--372, 2006.

\bibitem{29}
M.~Nazari, A.~Oroojlooy, L.~Snyder, and M.~Takac, ``Reinforcement learning for solving the vehicle routing problem," In \emph{Advances in Neural Information Processing Systems}, pp.~9839--9849, 2018.

\bibitem{30}
I.~Bello, H.~Pham, Q.~V.~Le, M.~Norouzi, and S.~Bengio, ``Neural combinatorial optimization with reinforcement learning," \emph{arXiv preprint arXiv:1611.09940}, 2016.

\bibitem{31}
L.~Panait and S.~Luke, ``Cooperative multi-agent learning: The state of the art," \emph{Autonomous Agents and Multi-Agent Systems}, vol.~11, no.~3, pp.~387--434, 2005.

\bibitem{32}
J.~Z.~Leibo et al., ``Multi-agent reinforcement learning in sequential social dilemmas," In \emph{Conference on Autonomous Agents and Multi-Agent Systems}, 2017.

\bibitem{33}
X.~Wang and T.~Sandholm, ``Reinforcement learning to play an optimal Nash equilibrium in team Markov games," In \emph{Advances in Neural Information Processing Systems}, pp.~1603--1610, 2003.

\bibitem{34}
J.~Peters and S.~Stefan, ``Natural actor-critic," \emph{Neurocomputing}, vol.~71, no~7--9, pp.~1180-1190, 2008.

\bibitem{35}
V.~R.~Konda and J.~N.~Tsitsiklis, ``Actor-critic algorithms," In \emph{Advances in Neural Information Processing Systems}, pp.~1008--1014, 2000.

\bibitem{36}
H.~He, J.~Boyd-Graber, K.~Kwok, and H.~Daume III, ``Opponent modeling in deep reinforcement learning," In \emph{International Conference on Machine Learning}, pp.~1804--1813, 2016.

\bibitem{37}
F.~Southey, M.~P.~Bowling, B.~Larson, C.~Piccione, N.~Burch, D.~Billings, and C.~Rayner, ``Bayes' bluff: Opponent modelling in poker," \emph{arXiv preprint arXiv:1207.1411}, 2012.

\bibitem{38}
I.~Goodfellow, J.~Pouget-Abadie, M.~Mirza, B.~Xu, D.~Warde-Farley, S.~Ozair, A.~Courville, and Y.~Bengio, ``Generative adversarial nets," In \emph{Advances in Neural Information Processing Systems}, pp.~2672--2680, 2014.

\bibitem{38b}
G.~Palmer et al., ``Lenient multi-agent deep reinforcement learning," In \emph{International Conference on Autonomous Agents and MultiAgent Systems}, 2018.

\bibitem{38c}
L.~Bu, B.~Robert, and D.S.~Bart, ``A comprehensive survey of multiagent reinforcement learning," \emph{IEEE Transactions on Systems, Man, and Cybernetics}, vol.~38, pp.~156--172, 2008.

\bibitem{38d}
K.~Tuyls and W.~Gerhar, ``Multiagent learning: Basics, challenges, and prospects," \emph{AI Magazine}, vol.~33, 2012.

\bibitem{39}
S.~Natarajan and P.~Tadepalli, ``Dynamic preferences in multi-criteria reinforcement learning," In \emph{International Conference on Machine Learning}, pp.~601--608, 2005.

\bibitem{40}
D.~M.~Roijers, P.~Vamplew, S.~Whiteson, and R.~Dazeley, ``A survey of multi-objective sequential decision-making," \emph{Journal of Artificial Intelligence Research}, vol.~48, pp.67--113, 2013.

\bibitem{41}
K.~Van Moffaert, and A.~Nowe, ``Multi-objective reinforcement learning using sets of pareto dominating policies," \emph{The Journal of Machine Learning Research}, vol.~15, no.~1, pp.~3483--3512, 2014.

\bibitem{41b}
L.~Barrett and S.~Narayanan, ``Learning all optimal policies with multiple criteria," In \emph{ International Conference on Machine Learning}, pp.~41--47, 2008.

\bibitem{42}
P.~Vamplew, R.~Dazeley, A.~Berry, R.~Issabekov, and E.~Dekker, ``Empirical evaluation methods for multiobjective reinforcement learning algorithms," \emph{Machine Learning}, vol.~84, no.~1--2, pp.~51--80, 2011.

\bibitem{43}
H.~Mossalam, Y.~M.~Assael, D.~M.~Roijers, and S.~Whiteson, ``Multi-objective deep reinforcement learning," \emph{arXiv preprint arXiv:1610.02707}, 2016.

\bibitem{44}
H.~Van Seijen, M.~Fatemi, J.~Romoff, R.~Laroche, T.~Barnes, and J.~Tsang, ``Hybrid reward architecture for reinforcement learning," In \emph{Advances in Neural Information Processing Systems}, pp.~5392--5402, 2017.

\bibitem{45}
T.~T.~Nguyen, ``A multi-objective deep reinforcement learning framework," \emph{arXiv preprint arXiv:1803.02965}, 2018.

\bibitem{45b}
S.~Shalev-Shwartz, S.~Shammah, and A.~Shashua, ``Safe, multi-agent, reinforcement learning for autonomous driving," \emph{arXiv preprint arXiv:1610.03295}, 2016.

\bibitem{45c}
A.~E.~Sallab, M.~Abdou, E.~Perot, and S.~Yogamani, ``Deep reinforcement learning framework for autonomous driving," \emph{Electronic Imaging}, pp.~70--76, 2017.

\bibitem{46}
D.~Amodei, C.~Olah, J.~Steinhardt, P.~Christiano, J.~Schulman, and D.~Mane. Concrete problems in AI safety. In \emph{arXiv:1606.06565}, 2016.

\bibitem{47}
P.~F.~Christiano, J.~Leike, T.~Brown, M.~Martic, S.~Legg, and D.~Amodei. Deep reinforcement learning from human preferences. In \emph{Advances in Neural Information Processing Systems}, pages 4302--4310, 2017.

\bibitem{48}
N.~D.~Nguyen, T.~T.~Nguyen, S.~Nahavandi, ``Multi-agent behavioral control system using deep reinforcement learning," \emph{Neurocomputing}, 2019.

\bibitem{109}
N.~D.~Nguyen and T.~T.~Nguyen ``Fruit-API," https://github.com/garlicdevs/Fruit-API, 2019.

\bibitem{49}
V.~Mnih et al., ``Playing atari with deep reinforcement learning," \emph{arXiv preprint arXiv:1312.5602}, 2013.

\bibitem{50}
H.~V.~Hasselt, G.~Arthur, and S.~David, ``Deep reinforcement learning with double q-learning," In \emph{Conference on Artificial Intelligence}, 2016.

\bibitem{51}
Z.~Wang et al., ``Dueling network architectures for deep reinforcement learning," \emph{arXiv preprint arXiv:1511.06581}, 2015.

\bibitem{52}
T.~Schaul et al., ``Prioritized experience replay," \emph{arXiv preprint arXiv:1511.05952}, 2015.

\bibitem{53}
M.~Hausknecht and S.~Peter, ``Deep recurrent q-learning for partially observable mdps," In \emph{AAAI Fall Symposium Series}, 2015.

\bibitem{54}
I.~Sorokin et al., ``Deep attention recurrent Q-network," \emph{arXiv preprint arXiv:1512.01693}, 2015.

\bibitem{55}
M.~Hessel, J.~Modayil, H.~Van Hasselt, T.~Schaul, G.~Ostrovski, W.~Dabney, D.~Horgan, B.~Piot, M.~Azar, and D.~Silver, ``Rainbow: Combining improvements in deep reinforcement learning," In \emph{AAAI Conference on Artificial Intelligence}, 2018.

\bibitem{56}
V.~Mnih et al., ``Asynchronous methods for deep reinforcement learning," \emph{International Conference on Machine Learning}, 2016.

\bibitem{57}
M.~Jaderberg et al., ``Reinforcement learning with unsupervised auxiliary tasks," \emph{arXiv preprint arXiv:1611.05397}, 2016.

\bibitem{58}
D.~Silver, G.~Lever, N.~Heess, T.~Degris, D.~Wierstra, and M.~Riedmiller, ``Deterministic policy gradient algorithms," in \emph{International Conference on Machine Learning}, 2014.

\bibitem{59}
T.P~Lillicrap, J.Hunt, A.~Pritzel, N.~Heess, T.~Erez, Y.~Tassa, D.~Silver, and D.~Wierstra, ``Continuous control with deep reinforcement learning," \emph{arXiv preprint arXiv:1509.02971}, 2015.

\bibitem{60}
R.~Lowe, Y.~Wu, A.~Tamar, J.~Harb, O.P.~Abbeel, and I.~Mordatch, ``Multi-agent actor-critic for mixed cooperative-competitive environments," In \emph{Advances in Neural Information Processing Systems}, pp.~6379--6390, 2017.

\bibitem{61}
J.~Schulman, S.~Levine, P.~Abbeel, M.~Jordan, and P.~Moritz, ``Trust region policy optimization," In \emph{International Conference on Machine Learning}, pp.~1889--1897, 2015.

\bibitem{62}
T.~Van Erven and P.~Harremos, ``Renyi divergence and Kullback-Leibler divergence," \emph{IEEE Transactions on Information Theory}, vol.~60, no.~7, pp.~3797--3820, 2014.

\bibitem{64}
Y.~Wu, E.~Mansimov, R.B.~Grosse, S.~Liao, and J.~Ba, ``Scalable trust-region method for deep reinforcement learning using kronecker-factored approximation," In \emph{Advances in Neural Information Processing Systems}, pp.~5279--5288, 2017.

\bibitem{65}
Z.~Wang, V.~Bapst, N.~Heess, V.~Mnih, R.~Munos, K.~Kavukcuoglu, and N.~de Freitas,  ``Sample efficient actor-critic with experience replay," \emph{arXiv preprint arXiv:1611.01224}, 2016.

\bibitem{63}
J.~Schulman, F.~Wolski, P.~Dhariwal, A.~Radford, and O.~Klimov, ``Proximal policy optimization algorithms," \emph{arXiv preprint arXiv:1707.06347}, 2017.

\bibitem{65b}
O.~Nachum, M.~Norouzi, K.~Xu, and D.~Schuurmans, ``Bridging the gap between value and policy based reinforcement learning," In \emph{Advances in Neural Information Processing Systems}, pp.~2775--2785, 2017.

\bibitem{65c}
B.~O'Donoghue, R.~Munos, K.~Kavukcuoglu, and V.~Mnih, ``Combining policy gradient and Q-learning," \emph{arXiv preprint arXiv:1611.01626}, 2016.

\bibitem{65d}
J.~Schulman, X.~Chen, and P.~Abbeel, ``Equivalence between policy gradients and soft q-learning," \emph{arXiv preprint arXiv:1704.06440}, 2017.

\bibitem{65e}
A.~Gruslys, W.~Dabney, M.G.~Azar, B.~Piot, M.~Bellemare, and R.~Munos, ``The Reactor: A fast and sample-efficient actor-critic agent for reinforcement learning," \emph{arXiv preprint arXiv:1704.04651}, 2017.

\bibitem{65f}
S.S~Gu, T.~Lillicrap, R.E.~Turner, RZ.~Ghahramani, B.~Schölkopf, and S.~Levine, ``Interpolated policy gradient: Merging on-policy and off-policy gradient estimation for deep reinforcement learning," In \emph{Advances in Neural Information Processing Systems}, pp.~3846--3855, 2017.

\bibitem{65g}
P.~S.~Castro, S.~Moitra, C.~Gelada, S.~Kumar, and M.~G.~Bellemare, ``Dopamine: A research framework for deep reinforcement learning," \emph{arXiv preprint arXiv:1812.06110}, 2018.

\bibitem{65h}
E.~Liang, R.~Liaw, R.~Nishihara, P.~Moritz, R.~Fox, J.~Gonzalez, and I.~Stoica, ``Ray rllib: A composable and scalable reinforcement learning library," \emph{arXiv preprint arXiv:1712.09381}, 2017.

\bibitem{65j}
C.~Hesse, M.~Plappert, A.~Radford, J.~Schulman, S.~Sidor, and Y.~Wu, ``OpenAI baselines," 2017.

\bibitem{65k}
S.~Tokui, K.~Oono, S.~Hido, and J.~Clayton, ``Chainer: a next-generation open source framework for deep learning," In \emph{Proceedings of Workshop on Machine Learning Systems in Conference on Neural Information Processing Systems}, pp.~1--6, 2015.

\bibitem{65n}
K.~Miyoshi, 2017. Available: https://github.com/miyosuda/unreal.

\bibitem{66}
A.~Tampuu \emph{et al.}, ``Multiagent cooperation and competition with deep reinforcement learning," \emph{PloS One}, vol.~12, no.~4, Apr. 2017.

\bibitem{67}
J.Z.~Leibo, v.~Zambaldi, M.~Lanctot, J.~Marecki, and T.~Graepel, ``Multi-agent reinforcement learning in sequential social dilemmas," In \emph{Conference on Autonomous Agents and MultiAgent Systems}, pp.~464--473, 2017.

\bibitem{68}
L.~Bu, B.~Robert, and D.S.~Bart, ``A comprehensive survey of multiagent reinforcement learning," \emph{IEEE Transactions on Systems, Man, and Cybernetics}, vol.~38, pp.~156--172, 2008.

\bibitem{69}
K.~Tuyls and W.~Gerhar, ``Multiagent learning: Basics, challenges, and prospects," \emph{AI Magazine}, vol.~33, 2012.

\bibitem{70}
G.~Palmer, K.~Tuyls, D.~Bloembergen, and R.~Savani, ``Lenient multi-agent deep reinforcement learning," In \emph{International Conference on Autonomous Agents and MultiAgent Systems}, pp.~443--451, 2018.

\bibitem{71}
L.~Kraemer and B.~Banerjee, ``Multi-agent reinforcement learning as a rehearsal for decentralized planning," \emph{Neurocomputing}, pp.~82--94, 2016.

\bibitem{72}
J.~Foerster, Y.~M.~Assael, N.~de~Freitas, and S.~Whiteson, ``Learning to communicate with deep multi-agent reinforcement learning," in \emph{Advances in Neural Information Processing Systems}, pp.~2137--2145, 2016.

\bibitem{73}
S.~Sukhbaatar and R.~Fergus, ``Learning multiagent communication with backpropagation," in \emph{Advances in Neural Information Processing Systems}, pp.~2244--2252, 2016.

\bibitem{74}
J.~K.~Gupta, M.~Egorov, and M.~Kochenderfer, ``Cooperative multi-agent control using deep reinforcement learning," In \emph{International Conference on Autonomous Agents and Multiagent Systems}, pp.~66--83, 2017.

\bibitem{75}
T.~Nguyen, N.~D.~Nguyen, and S.~Nahavandi, ``Deep reinforcement learning for multi-agent systems: A review of challenges, solutions and applications," \emph{arXiv preprint arXiv:1812.11794}, 2018.

\bibitem{76}
M.~Egorov, ``Multi-agent deep reinforcement learning," \emph{CS231n: Convolutional Neural Networks for Visual Recognition}, 2016.

\bibitem{77}
L.~Bu, B.~Robert, and D.S.~Bart, ``A comprehensive survey of multiagent reinforcement learning," \emph{IEEE Transactions on Systems, Man, and Cybernetics}, vol.~38, pp.~156--172, 2008.

\bibitem{78}
R.~Lowe, Y.~Wu, A.~Tamar, J.~Harb, O.P.~Abbeel, and I.~Mordatch, ``Multi-agent actor-critic for mixed cooperative-competitive environments," In \emph{Advances in Neural Information Processing Systems}, pp.~6379--6390, 2017.

\bibitem{79}
F.~Girosi, M.~Jones, and T.~Poggio, ``Regularization theory and neural networks architectures," \emph{Neural Comput.}, vol.~7, no.~2, pp.~219--269, 1995.

\bibitem{80}
I.~J.~Goodfellow, M.~Mirza, D.~Xiao, A.~Courville, and Y.~Bengio, ``An empirical investigation of catastrophic forgetting in gradient-based neural networks," \emph{arXiv:1312.6211 [cs, stat]}, Dec. 2013.

\bibitem{81}
S.~Thrun and L.~Pratt, \emph{Learning to Learn}. Boston, MA: Kluwer Academic Publishers, 1998.

\bibitem{82}
A.~A.~Rusu \emph{et al.}, ``Progressive neural networks," \emph{arXiv:1606.04671 [cs]}, 2016.

\bibitem{86}
J.~Kirkpatrick \emph{et al.}, ``Overcoming catastrophic forgetting in neural networks," in \emph{Proc. Nat. Acad. Sci.}, pp.~3521--3526, 2017.

\bibitem{86b}
C.~Fernando \emph{et al.}, ``Pathnet: Evolution channels gradient descent in super neural networks," \emph{arXiv preprint arXiv:1701.087}, 2017.

\bibitem{83}
A.~A.~Rusu \emph{et al.}, ``Policy distillation," \emph{arXiv:1511.06295 [cs]}, Nov.~2015.

\bibitem{84}
H.~Yin and S.~J.~Pan, ``Knowledge transfer for deep reinforcement learning with hierarchical experience replay," in \emph{Proc. AAAI Conf. Artif. Intell.}, pp.~1640--1646, Jan. 2017.

\bibitem{85}
E.~Parisotto, J.~L.~Ba, and R.~Salakhutdinov, ``Actor-mimic: Deep multitask and transfer reinforcement learning," \emph{arXiv:1511.06342 [cs]}, 2015.

\bibitem{87}
H.~Yin and S.~J.~Pan, ``Knowledge transfer for deep reinforcement learning with hierarchical experience replay," in \emph{Proc. AAAI Conf. Artif. Intell.}, pp.~1640--1646, Jan. 2017.

\bibitem{88}
M.~Wulfmeier, I.~Posner, and P.~Abbeel, ``Mutual alignment transfer learning," \emph{arXiv preprint arXiv:1707.07907}, 2017.

\bibitem{89}
M.~Grzes, and D.~Kudenko, ``Online learning of shaping rewards in reinforcement learning," \emph{Neural Networks}, vol.~23, no.~4, pp.~541--550, 2010.

\bibitem{90}
A.~G.~Barto and S.~Mahadevan, ``Recent advances in hierarchical reinforcement learning," \emph{Discrete Event Dyn. Syst.}, vol.~13, no.~4, pp.~341--379, 2003.

\bibitem{91}
T.~D.~.Kulkarni, K.~R.~Narasimhan, A.~Saeedi, and J.~B.~Tenenbaum, ``Hierarchical deep reinforcement learning: Integrating temporal abstraction and intrinsic motivation," in \emph{Adv. Neural Inf. Process. Syst.}, pp.~3675--3683, 2016.

\bibitem{92}
Y.~Burda, H.~Edwards, D.~Pathak, A.~Storkey, T.~Darrell, and A.A.~Efros, ``Large-scale study of curiosity-driven learning," \emph{arXiv preprint arXiv:1808.04355}, 2018.

\bibitem{93}
D.~Pathak, P.~Agrawal, A.A.~Efros, and T.~Darrell, ``Curiosity-driven exploration by self-supervised prediction," In \emph{Conference on Computer Vision and Pattern Recognition Workshops}, pp.~16--17, 2017.

\bibitem{94}
G.~Ostrovski, M.~G.~Bellemare, A.~van den Oord, and R.~Munos, ``Count-based exploration with neural density models," In \emph{International Conference on Machine Learning}, pp.~2721--2730, 2017.

\bibitem{95}
M.~Andrychowicz \emph{et al.}, ``Hindsight experience replay," In \emph{Advances in Neural Information Processing Systems}, 2017.

\bibitem{96}
Y.~Bengio, J.~Louradour, R.~Collobert, and J.~Weston, ``Curriculum learning," In \emph{International Conference on Machine Learning}, pp. 41--48, 2009.

\bibitem{97}
A.~Santoro, R.~Faulkner, D.~Raposo, J.~Rae, M.~Chrzanowski, T.~Weber, D.~Wierstra, O.~Vinyals, R.~Pascanu, and T.~Lillicrap, ``Relational recurrent neural networks," In \emph{Advances in Neural Information Processing Systems}, pp.~7299--7310, 2018.

\bibitem{98}
E.~Parisotto and R.~Salakhutdinov, ``Neural map: Structured memory for deep reinforcement learning," \emph{arXiv preprint arXiv:1702.08360}, 2017.

\bibitem{99}
D.~Horgan, J.~Quan, D.~Budden, G~Barth-Maron, M.~Hessel, H.~Van Hasselt, and D.~Silver, ``Distributed prioritized experience replay," \emph{arXiv preprint arXiv:1803.00933}, 2018.

\bibitem{100}
A.~Stooke and P.~Abbeel, ``Accelerated methods for deep reinforcement learning," \emph{arXiv preprint arXiv:1803.02811}, 2018.

\balance

\bibitem{101}
E.~Liang, R.~Liaw, P.~Moritz, R.~Nishihara, R.~Fox, K.~Goldberg, and I.~Stoica, ``Rllib: Abstractions for distributed reinforcement learning," \emph{arXiv preprint arXiv:1712.09381}, 2017.

\bibitem{102}
J.~Ho and S.~Ermon, ``Generative adversarial imitation learning," In \emph{Advances in Neural Information Processing Systems}, pp.~4565--4573, 2016.

\bibitem{103}
M.~G.~Bellemare, Y.~Naddaf, J.~Veness, and M.~Bowling, ``The arcade learning environment: An evaluation platform for general agents," \emph{J. Artif. Intell. Res.}, vol.~47, pp.~253--279, 2013.

\bibitem{104}
G.~Brockman \emph{et al.}, ``OpenAI gym," \emph{arXiv:1606.01540 [cs]}, 2016.

\bibitem{105}
E.~Todorov, T.~Erez, and Y.~Tassa, ``Mujoco: A physics engine for model-based control," In \emph{International Conference on Intelligent Robots and Systems}, pp.~5026--5033, 2012.

\bibitem{106}
M.~Plappert, ``keras-rl," https://github.com/matthiasplappert/keras-rl, 2016.

\bibitem{108}
Y.~Duan, X.~Chen, R.~Houthooft, J.~Schulman, and P.~Abbeel, ``Benchmarking deep reinforcement learning for continuous control," In \emph{International Conference on Machine Learning}, pp.~1329--1338, 2016.

\bibitem{107}
M.~Schaarschmidt, A.~Kuhnle, and K.~Fricke, ``TensorForce: A TensorFlow library for applied reinforcement learning," 2017.

\bibitem{107b} A.~Paszke \emph{et al.}, ``PyTorch: An imperative style, high-performance deep learning library," In \emph{Advances in Neural Information Processing Systems}, 2019.

\bibitem{107c} B.~Ellis, J.~Stylos, and B.~Myers, ``The factory pattern in API design: A usability evaluation," In \emph{International Conference on Software Engineering}, 2007.

\bibitem{108}
N.~D.~Nguyen and T.~T.~Nguyen, ``FruitLAB," http://fruitlab.org/, 2019.


\end{thebibliography}
\end{document}